# ADAPTIVE SPECULAR REFLECTION DETECTION AND INPAINTING IN COLONOSCOPY VIDEO FRAMES

*Mojtaba Akbari[1], Majid Mohrekesh[1], S.M.Reza Soroushmehr[2,3],*
*Nader Karimi[1], Shadrokh Samavi[1,3], Kayvan Najarian[2,3]*

[1]Dept. of Elect. and Comp. Eng., Isfahan University of Technology, Isfahan 84156-83111, Iran
[2]Dept. of Computational Medicine and Bioinformatics, University of Michigan, Ann Arbor, MI, U.S.A.
[3]Michigan Center for Integrative Research in Critical Care, University of Michigan, Ann Arbor, MI, U.S.A.

## ABSTRACT

Colonoscopy video frames might be contaminated by bright spots with unsaturated values known as specular reflection. Detection and removal of such reflections could enhance the quality of colonoscopy images and facilitate diagnosis procedure. In this paper we propose a novel two-phase method for this purpose, consisting of detection and removal phases. In the detection phase, we employ both HSV and RGB color space information for segmentation of specular reflections. We first train a non-linear SVM for selecting a color space based on image statistical features extracted from each channel of the color spaces. Then, a cost function for detection of specular reflections is introduced. In the removal phase, we propose a two-step inpainting method which consists of appropriate replacement patch selection and removal of the blockiness effects. The proposed method is evaluated by testing on an available colonoscopy image database where accuracy and Dice score of 99.68% and 71.79% are achieved respectively.

## 1. INTRODUCTION

Specular reflection refers to bright pixels/patches with unsaturated colors which could occur in medical imaging. This type of reflection is a common cause of degradation in quality of colonoscopy and endoscopy images. Physicians usually prefer elimination of specular reflection in medical images [1]. In the case of endoscopy or colonoscopy, the imaging device carries flash lights that produce reflections in presence of moisture in the inner surface of gastrointestinal tract. Detection of these reflections is a challenging task due to their shape and color variations in different images. In order to remove these specular reflections, image inpainting can be employed that fills these patches with similar shapes and textures to textures of the image in that region.

Different color spaces have been used for detection of specular reflections. Hue-Saturation-Value (HSV) color space is very informative in terms of detection of specular reflections. One of the characteristics of reflections is that colors are unsaturated and the *value* (V) component is saturated. Oh *et al.* [2] first convert an image from RGB to HSV and then using a thresholding approach and texture/color information they detect bright regions. The method proposed in [3] also uses HSV color space and thresholding on the $S$ and V channels in the first stage of detection. Then they combine this data with that of "wavelet transform module maxima" which is proposed in [4] to improve the detection performance. Algorithm of [5] uses hysteresis thresholds on the $S$ and $V$ channels and updates threshold values for better detection.

A method proposed in [6] that uses the $Y$ channel of the YUV color space to remove the specular reflection. Method of [7] uses the $Y$ channel in CIE-XYZ color space and compares the $Y$ channel of the input image with its normal version to detect specular reflections. Guo *et al.* proposed a method based upon thresholding on grayscale image to segment specular reflections [8].

Due to low signal to noise ratio in occurrence of specular reflections, the original information of the organism, which is covered by the reflection, is not recoverable in most cases. Therefore, removal of such reflections needs an inpainting method to repair the reflection after its detection. Review of inpainting algorithms is available in [9] that categorizes inpainting algorithms into different groups based on their mechanisms.

In this paper we propose a novel reflection detection method based on both RGB and HSV color spaces with an SVM classifier. We also introduce our proposed inpainting method based on patch selection around each reflection region. Moreover, we propose an edge smoothing algorithm to enhance the quality of inpainted image.

The structure of the paper is as follows. Section 2 is dedicated to the proposed reflection detection. In Section 3 we describe the proposed inpainting method and then in section 4 we evaluate our algorithm on the CVC-ColonDB database. Concluding remarks are offered in section 5.

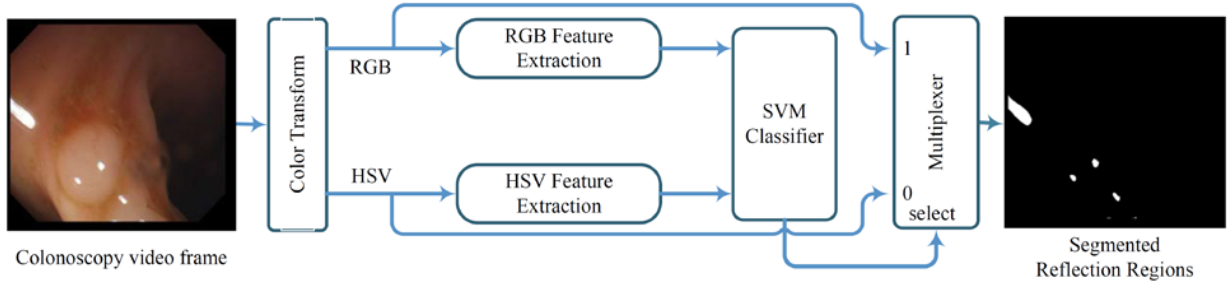

Figure 1. Block diagram of the proposed reflection detection method.

## 2. PROPOSED DETECTION METHOD

In this section we propose our novel reflection detection method. The proposed method selects appropriate color space based on image features to detect specular reflections. Fig. 1 shows general concept of the proposed reflection detection method. The overall function of the method is an adaptive detection of reflection based on the analysis that shows which of RGB or HSV color spaces can better detect the reflections of each specific image. Details of what is happening in this block diagram are in the following sub-sections.

### 2.1. Proposed RGB Detection Method

RGB is the most well-known color space in image processing algorithms. RGB color space is near to human perceptual system. Because of brightness of specular reflection, we built our RGB detection method based on thresholding on all three channels in this color space and voting scheme to label each pixel. Our voting scheme labels each pixel as a specular reflection if two out of three channels address that pixel as a reflection.

### 2.2. Proposed HSV Detection Method

HSV is one of the most useful color spaces in image processing because of its capability to separate the image into uncorrelated image components with different specificities. Fig. 2 shows an input RGB image of the colonoscopy and its corresponding HSV channels.

Fig. 2 shows that both the $S$ and $V$ channels have good indicators of reflection while the $H$ channel does not have any information in this regard. The $S$ channel has low values in reflection areas because of rather similar color saturations in such regions. From the color transform point of view, reflections could be considered as aggregation of all colors in one pixel that restricts the higher power of one color and subsequently restricts the saturation of that color and also its value. Hence, a function of $(1-S) \times V$ is used as the segmentation criterion for the reflections. The term $(1-S)$ causes higher values for reflections from the saturation point of view and multiplication of the two factors results in direct effect of changes in both terms. Our experiments show that this function works well in finding large reflections and needs to be modified to find small size reflections.

Fig. 3 demonstrates the graphical path of achieving the final cost function through images. Fig. 3 (b) shows the result of utilizing $(1-S) \times V$ that is rather appropriate for large reflections and not efficient for small ones. On the other hand, even though hue values are high at some points, there are some points inside the reflection areas with low hue values. Fig. 2 (b) also shows that hue values of the pixels in boundaries of reflections highly fluctuate in the small neighborhoods. In addition, the fluctuations of hue are increased specially when the surface of each reflection reduces and the facing boundaries consequently approach to each other. This means that the fluctuations of hue are high although the value of hue is not very high or very low in the range of hue values. Another phenomenon is that the high amount of reflection means a very bright semi-white color which could be achieved from any path in the HSV model. A small deviation from white to any other pure color would result in a large change in the hue of the pixel. Hence, we calculate local variance of hue in sliding windows of size 3×3 to take local variance of hue into account. Aggregation of the effects of $(1-S) \times V$ with that of $variance(H)$ is shown in Fig. 3 (d). In this figure we can see higher precision in finding the reflection regions.

Based on the above analysis, Fig. 4 shows a block diagram of the proposed method in HSV color space which contains the following steps:
1. Transformation from RGB to HSV color space.
2. Analyzing each $H$, $S$ and $V$ channels separately to generate three matrices called $\boldsymbol{H}$, $\boldsymbol{\Sigma}$ and $\boldsymbol{Y}$ respectively. Details of properties of each channel will be discussed later.
3. Aggregating all features of three channels to produce the proposed cost function in (1).
4. Thresholding on the result, using a statistically constant threshold function as proposed in (2).

$$C_{HSV} = \boldsymbol{H} + \boldsymbol{\Sigma} \times \boldsymbol{Y} \quad (1)$$
$$t_0 = \mu + k \times \sigma \quad (2)$$

The block diagram of Fig. 4 contains some minor refinements as compared to the above mentioned four steps. These refinements are empirically proved to enhance the efficiency of the algorithm. The details in processing of each channel are as follows.

- *H Channel*

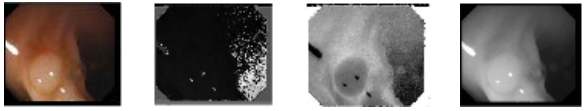

(a)　　　　(b)　　　　(c)　　　　(d)

Figure 2. HSV color space, (a) Original image, (b) H, (c) S and (c) V channels of the original image

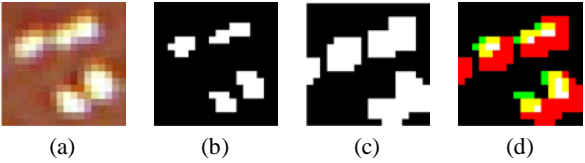

(a)　　　　(b)　　　　(c)　　　　(d)

Figure 3. (a) Original Reflected Region, (b) output of $(1 - S) \times V$, (c) output of $variance(H)$, (d) RGB result, Red: $variance(H)$, Green: $(1 - S) \times V$, Blue: Ground truth

1. Extracting a unique 3×3 patch for each pixel
2. Calculating the variance of patch to represent the local variance of central pixel of the patch.
3. Normalization of the local variances for uniform performance of a cost function.

- **S Channel**
1. Normalization of (-S) and then a unit shift.
2. Applying a ramp function to eliminate negative values, this means maintaining just lower values of saturation that have higher probability of bright reflections.

- **V Channel**
1. Normalization of V and then a unit shift.
2. Applying a ramp function to eliminate negative values. This means maintaining just higher values of V that have higher probability of bright reflections.

Hence, a threshold of the form $\mu + k\sigma$ performs a uniform thresholding on the dataset, which is adaptive to the statistics of each image. The higher values of $k$ result in segmenting smaller regions as reflection. Values more than this threshold are considered as reflections and others are considered as normal pixels.

### 2.3. Selecting Color Space with SVM

We use SVM as a classifier to select appropriate color space based on input images. Non-linear SVM is the improved version of soft margin SVM that can separate two classes with non-linear boundaries. In our method we use non-linear SVM with Gaussian kernel for selecting appropriate color space. We use binary labeling for training of non-linear SVM classifier and evaluate both RGB and HSV methods in our database to decide which one outperforms the other one.

The non-linear SVM is trained with 12 statistical features including mean and standard deviation of each channel of RGB and HSV color spaces. We normalize all input features individually with respect to their mean and standard deviation in whole training dataset in order to improve our training process.

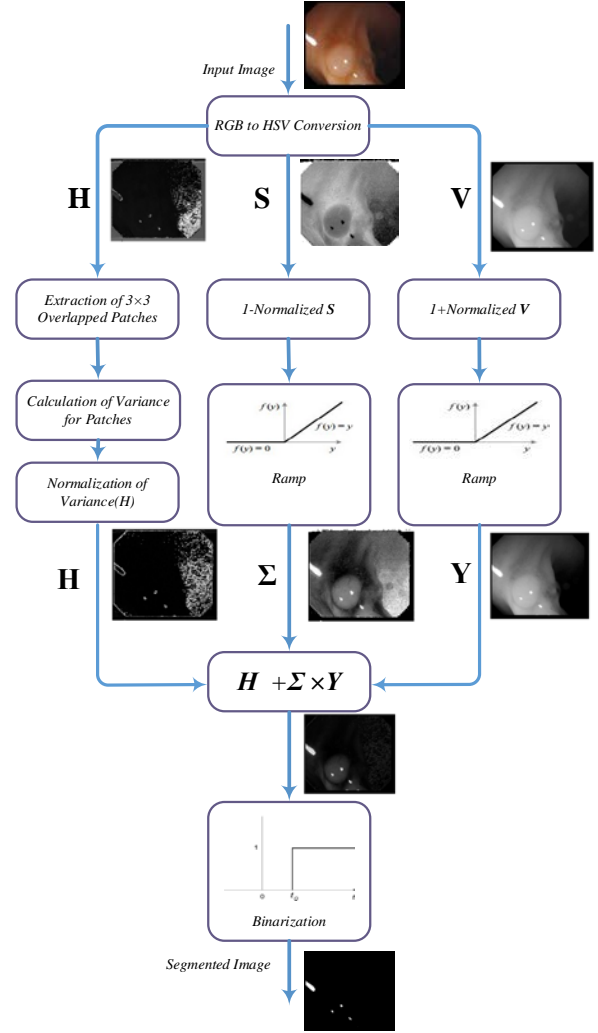

Figure 4. Block diagram of the proposed reflection detection method using HSV color space.

### 3. PROPOSED INPAINTING METHOD

Our proposed inpainting method is based on our previous work [10]. The method proposed in [10] employs a cost function for selecting patches around each connected component in four directions. Here, we add a correlation term to the cost function for selecting patches more similar to main reflection patch. Adding the correlation term will improve our algorithm performance. The improved cost function for selecting candidate patches around each reflecting connected components is shown in (3).

$$Cost = \Delta_\mu \times \Delta_\sigma \times d \times (1 - NC) \qquad (3)$$

As there might be artifacts on edges of the segmented regions, we apply a smoothing method based on gradual vanishing of edge artifacts alongside the edges.

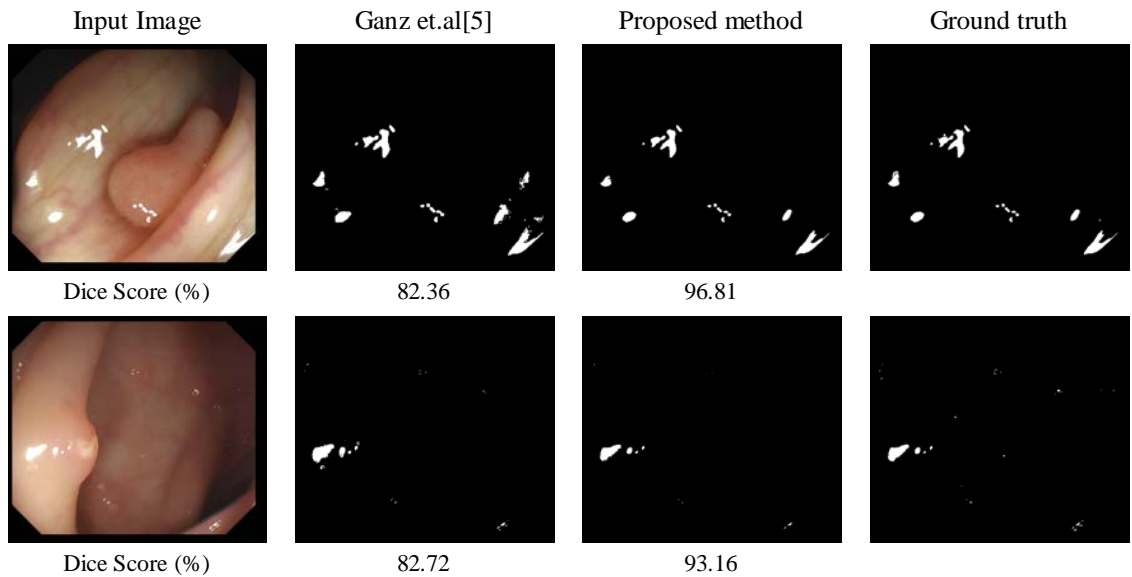

Figure 5. Visual and Dice score comparison of the proposed method with the method of [5].

Table I. Results of the proposed method and comparison with the method of [5].

|  | Dice (%) | Accuracy (%) | Specificity (%) | Precision (%) |
|---|---|---|---|---|
| **RGB Method** | 60.26 | 99.62 | **99.96** | **87.52** |
| **HSV Method** | 67.34 | 99.58 | 99.78 | 66.44 |
| **Proposed Method** | **71.79** | **99.68** | 99.92 | 82.78 |
| **Ganz et al [5]** | 61.76 | 99.34 | 99.43 | 48.67 |

The starting point is a two-pixel width ring around the reflection, inner pixel from the substitute replaced region and the outer from outer layer of original reflection patch.

For each pixel, a 3×3 patch is selected and the value of intensity is changed to a random value with a normal distribution. The mean and standard deviation of this distribution are respectively equal to the patch mean and patch standard deviation for all three channels of RGB. After replacing all pixels of this two-pixel ring, we dilate the ring just by one pixel and repeat the above process for this new ring. The number of dilations and set of standard deviations for random normal values depend on the image type which will be mentioned in experimental results for this specific type of image.

## 4. EXPERIMENTAL RESULTS

We evaluate our proposed reflection detection and inpainting method using CVC-ColonDB database. The database is available for the current challenge on polyp detection in colonoscopy videos [11]. This database contains colonoscopy videos from 15 video sequences. Our database has ground truth for colonic polyps and we randomly select 100 images from whole dataset to evaluate our proposed reflection detection and inpainting method. We manually segment specular reflections for these 100 images. All methods are implemented on Corei7 Intel CPU and 6GB of RAM memory and MATLAB R2014a translator. Eight iterations of edge vanishing are applied on the boundary pixels with coefficients of standard deviation: 0.71, 0.99, 1.2, 0.66, 0.66, 0.66, 0.66 and 0.74 for random normal values. We empirically use the interval (4.45 , 4.55) as the best interval for values of $k$ and consequently, the best threshold value to extract the 0.6% of reflecting pixels near $\mu + 4.5\sigma$.

We also evaluate our proposed method for all images in our dataset with a number of quality assessment criteria in image processing. Results of the proposed method are reported in Table I. We also implement reflection detection of [5] and its results are shown in Table I. Fig. 5 also shows image results of the proposed method in comparison with the method proposed in [5].

## 5. CONCLUSION

We proposed a novel adaptive specular reflection detection method in the colonoscopy frames. The detection method was based on adaptation between the proposed RGB and HSV detection methods. The adaptation was performed using a non-linear SVM classifier and the resulting detected reflection was inpainted by our inpainting method. The experimental results show better detection of reflection in comparison with similar research works in this field. The inpainting was performed to prepare the images for tasks, such as polyp detection, that may later on be performed on these frames.